# TinyBEV: Cross-Modal Knowledge Distillation for Efficient Multi-Task Bird's-Eye-View Perception and Planning

Reeshad Khan
University of Arkansas
Fayetteville, Arkansas
rk010@uark.edu

John Gauch
University of Arkansas
Fayetteville, Arkansas
jgauch@uark.edu

## Abstract

*We present* TinyBEV, *a unified, camera-only Bird's-Eye-View (BEV) framework that distills the full-stack capabilities of a large planning-oriented teacher (UniAD [19]) into a compact, real-time student model. Unlike prior efficient camera-only baselines such as VAD[23] and VADv2[7], TinyBEV supports the complete autonomy stack—3D detection, HD-map segmentation, motion forecasting, occupancy prediction, and goal-directed planning—within a streamlined 28M-parameter backbone, achieving a 78% reduction in parameters over UniAD [19]. Our model-agnostic, multi-stage distillation strategy combines feature-level, output-level, and adaptive region-aware supervision to effectively transfer high-capacity multi-modal knowledge to a lightweight BEV representation. On nuScenes[4], Tiny-BEV achieves 39.0 mAP for detection, 1.08 minADE for motion forecasting, and a 0.32 collision rate, while running 5× faster (11 FPS) and requiring only camera input. These results demonstrate that full-stack driving intelligence can be retained in resource-constrained settings, bridging the gap between large-scale, multi-modal perception-planning models and deployment-ready real-time autonomy.*

## 1. Introduction

Modern autonomous driving systems must perceive, predict, and plan within tight latency constraints to ensure safe navigation in dynamic environments [2, 17, 21]. End-to-end frameworks such as UniAD [19] have demonstrated that jointly optimizing detection, mapping, motion forecasting, and planning under a unified bird's-eye-view (BEV) representation can yield strong performance gains. UniAD [19]'s transformer-based architecture executes all tasks in a single forward pass on the nuScenes benchmark [4], but its large computational footprint (> $10^{12}$ FLOPs per frame) and reliance on dense multi-frame temporal fusion hinder real-time deployment on embedded ECUs and other resource-constrained platforms [1, 11, 36].

As noted in UniAD [19]'s original study, such comprehensive multi-task reasoning requires substantial compute and memory, motivating the need for lightweight alternatives that retain integrated decision-making capabilities [18, 43]. While research has explored enhancements such as depth estimation or behavioral prediction, the challenge remains to design compact architectures that maintain full-stack autonomy [3, 5, 12, 32].

Recent works address parts of this challenge. BEV-Former [31, 49], BEVDepth [15, 28, 29], and QD-BEV [55] advance monocular BEV perception but remain perception-centric and often depend on extensive temporal context [30, 40]. Multi-modal fusion methods such as BEVFusion [9, 35], FUTR3D [46], and OCFusion [51] achieve strong results by leveraging LiDAR–camera inputs, but require multi-sensor synchronization and additional hardware [44, 56]. On the compression side, methods like MMDistill [27], LabelDistill [24], and SCKD [48] distill single-task perception models, but do not preserve the integrated planning capability of full-stack systems like UniAD [19]. Camera-only baselines such as VAD[23] and VADv2[7] achieve real-time perception but omit forecasting and planning, limiting their applicability in decision-critical scenarios.

**TinyBEV: Real-Time Full-Stack Autonomy.** We propose **TinyBEV**, a unified *camera-only* student that distills the full-stack capabilities of the multi-modal, planning-oriented UniAD [19] into a compact architecture optimized for embedded deployment. TinyBEV performs 3D detection, semantic map segmentation, motion forecasting, occupancy prediction, and goal-directed planning within a single BEV encoder–decoder pipeline [6, 8, 50]. It achieves this through:

- *Multi-stage, cross-task distillation* — transferring feature-, output-, and behavior-level knowledge from the teacher across all tasks.
- *Adaptive region-aware learning* — emphasizing safety-critical regions (e.g., dynamic agents, drivable boundaries) with spatially weighted losses [10, 14].

| Method | Tasks | RT | Full |
|---|---|---|---|
| FCOS3D [45] | D | ✓ | ✗ |
| CenterPoint [52] | D + T | ✓ | ✗ |
| BEVFormer [31, 49] | D + M | ✗ | ✗ |
| FUTR3D [46] | D + F | ✗ | ✗ |
| UniAD [19] | D + M + F + P | ✗ | ✓ |
| MMDistill [27] | D | ✓ | ✗ |
| **TinyBEV (Ours)** | D + M + F + P | ✓ | ✓ |

Table 1. Contemporary autonomous driving systems. Abbreviations: D = Detection, T = Tracking, M = Mapping, F = Forecasting, P = Planning, RT = Real-time, Full = Full-stack.

- *Parameter-efficient BEV heads* — sharing features across all tasks without point cloud inputs, reducing model size and FLOPs [13, 34, 42].

On nuScenes, TinyBEV matches its teacher's performance in detection (mAP), forecasting (minADE), and planning (L2 error and collision rate) within small margins, while running 5× faster on RTX 4090 and 4× faster on Orin NX. As summarized in Table 1, TinyBEV uniquely combines *full-stack reasoning* with *real-time performance*, bridging the gap between unified perception–prediction–planning pipelines and practical on-vehicle deployment.

**Dataset and Teacher Selection.** Following our planning-oriented teacher model UniAD [19], we adopt the nuScenes dataset [4] as our sole evaluation benchmark. This ensures identical task supervision and direct comparability between teacher and student. We did not incorporate additional datasets or alternative teacher models in this study to avoid introducing variables unrelated to distillation efficacy. Future work will extend TinyBEV to other planning-capable teachers (e.g., VADv2[7]) and diverse datasets (e.g., Argoverse, Waymo) to evaluate generalization beyond the current setting.

## 2. Methodology

### 2.1. Background: UniAD [19] and Limitations

Contemporary autonomous driving systems have increasingly adopted unified architectures for jointly tackling perception, prediction, and planning tasks [1, 2, 11, 17, 21]. One prominent design is UniAD [19], which integrates these components into a holistic transformer-based framework using a Bird's-Eye-View (BEV) representation. This shared BEV allows for effective spatial-temporal reasoning across multiple modalities—camera, LiDAR, and HD map—achieving strong results on benchmarks such as nuScenes [4].

Despite its performance, UniAD [19] is computationally expensive. Its full-stack modeling involves heavy encoders, attention blocks, and dense temporal fusion, resulting in over 100M parameters and sub-real-time performance even on high-end GPUs. The authors themselves acknowledge the difficulty of coordinating such a comprehensive system under compute constraints and highlight the need for lightweight full-stack alternatives [18, 19, 32, 43].

### 2.2. Motivation for TinyBEV

To address these limitations, we propose **TinyBEV**—a compact BEV-centric student model distilled from the UniAD [19] teacher. The goal is to retain key full-stack capabilities (3D detection, motion forecasting, and planning) while drastically reducing computation and memory footprint [18, 34, 39, 42, 43, 50]. TinyBEV is tailored for deployment on real-time hardware such as ADAS ECUs and Orin-NX edge devices, achieving an order-of-magnitude speedup without large drops in accuracy [13, 34]. Following the Planning-oriented Autonomous Driving [19] baseline, we adopt the nuScenes dataset for training and evaluation to ensure comparability with the teacher model. Exploring alternative datasets and teacher models is left as future work.

### 2.3. Proposed Architecture: TinyBEV

Our student model adopts a simplified Lift-Splat-Shoot (LSS) [39] style projection mechanism that transforms image features into a planar BEV grid. Given multi-view images $I_c$, lightweight image backbones $\mathsf{B}$ extract features $F_c$:

$$F_c = \mathsf{B}(I_c), \quad c \in \{1, \ldots, C\} \quad (1)$$

These features are lifted to 3D via predicted depth distributions and pooled onto a common BEV grid $\mathbf{B}_S$:

$$\mathbf{B}_S = \text{LSS } \{F_c\}_{c=1}^{C} \quad (2)$$

Unlike UniAD [19], we avoid costly 4D temporal transformers and instead use shared BEV features $\mathbf{B}_S$ as input to all task heads, enabling efficient parameter reuse [26, 28, 31, 49].

The overall TinyBEV pipeline and distillation pathways are illustrated in Fig. 1.

### 2.4. Task-specific Heads

We employ lightweight convolutional and MLP modules for each task [31, 34, 39, 42, 49, 50]:

**Detection Head:** A convolutional decoder predicts 3D object centers, sizes, orientations, and velocities:

$$y_{\text{det}} = \sigma(\text{Conv}(\mathbf{B}_S)) \quad (3)$$

**Motion Forecasting Head:** We condition future trajectory predictions on detected agent states and BEV context:

$$\{\Delta \mathbf{x}_t\}_{t=1}^{T_f} = \mathsf{M}(\mathbf{X}_{\text{det}}, \mathbf{B}_S) \quad (4)$$

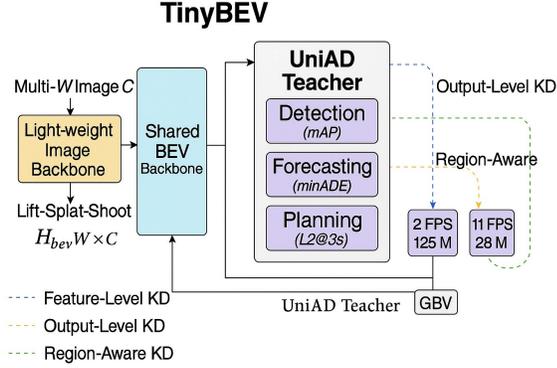

Figure 1. **Proposed TinyBEV architecture and multi-stage distillation pipeline.** Multi-view RGB images are processed by a lightweight ResNet-18 backbone and Lift–Splat–Shoot (LSS) projection to produce a shared BEV feature map (e.g., $128{\times}128{\times}C$). These features feed task-specific heads for 3D detection, motion forecasting, and planning (Eqs. 3, 4, 5). Distillation from the UniAD [19] teacher occurs at three levels: *feature-level* (solid blue, Eq. 6), *output-level* (dashed orange, Eqs. 7, 8, 9, 10), and *region-aware* (dotted red, Eq. 11), all contributing to the total objective in Eq. 12. Inset shows inference speed and parameter count (TinyBEV: 11 FPS / 28M; UniAD [19]: 2 FPS / 125M).

**Planning Head:** A compact MLP predicts future ego-vehicle trajectories from ego-centric BEV tokens:

$$\tau = \text{MLP}(\mathbf{B}_{\text{ego}}) \quad (5)$$

### 2.5. Cross-modal and Cross-task Knowledge Distillation

We transfer holistic knowledge from UniAD [19] through a three-stage distillation pipeline [8, 24, 27, 48, 50]:

**Feature-level Distillation:** Match BEV features between student and teacher:

$$\mathcal{L}_{\text{feat-KD}} = \frac{1}{N}\sum_{i=1}^{N} \|\mathbf{B}_{S,i} - \mathbf{B}_{T,i}\|_2^2 \quad (6)$$

**Detection and Regression Distillation:** Align heatmaps using KL-divergence and boxes via L1 loss:

$$\mathcal{L}_{\text{det-KD}} = \sum_u P_T(u)\log\frac{P_T(u)}{P_S(u)} \quad (7)$$

$$\mathcal{L}_{\text{bbox-KD}} = \sum_u \|\mathbf{b}_T(u) - \mathbf{b}_S(u)\|_1 \quad (8)$$

**Trajectory and Planning Distillation:** Use L2 regression losses to guide both agent trajectories and ego plans [3, 5, 12, 32]:

$$\mathcal{L}_{\text{mot-KD}} = \frac{1}{A}\sum_{a=1}^{A}\sum_{t=1}^{T_f} \|\hat{\mathbf{x}}_{a,t}^S - \hat{\mathbf{x}}_{a,t}^T\|_2^2 \quad (9)$$

$$\mathcal{L}_{\text{plan-KD}} = \sum_{t=1}^{T_f} \|\tau_t^S - \tau_t^T\|_2^2 \quad (10)$$

### 2.6. Adaptive Region-aware Distillation

To prioritize safety-critical regions (e.g., drivable lanes, dynamic agents), we apply attention-masked distillation [10, 14]:

$$\mathcal{L}_{\text{adaptive-KD}} = \frac{1}{|\mathsf{F}|}\sum_{f\in\mathsf{F}}\|\mathbf{B}_{S,f} - \mathbf{B}_{T,f}\|_2^2 \quad (11)$$

where $\mathsf{F}$ indexes salient BEV pillars defined by lane maps or foreground masks.

### 2.7. Overall Objective

We optimize a combined loss:

$$\begin{aligned}\mathcal{L}_{\text{total}} = {} & \mathcal{L}_{\text{GT}} + \lambda_{\text{feat}}\mathcal{L}_{\text{feat-KD}} + \lambda_{\text{det}}\mathcal{L}_{\text{det-KD}} \\ & + \lambda_{\text{mot}}\mathcal{L}_{\text{mot-KD}} + \lambda_{\text{plan}}\mathcal{L}_{\text{plan-KD}} + \lambda_{\text{adapt}}\mathcal{L}_{\text{adaptive-KD}}\end{aligned} \quad (12)$$

This pipeline enables TinyBEV to inherit the key perception, prediction, and planning capabilities of UniAD [19] at a fraction of the computational cost. By following the same dataset and evaluation setup as the teacher, we ensure fair comparison and reproducibility.

## 3. Experiments

### 3.1. Dataset and Setup

We evaluate **TinyBEV** on the nuScenes dataset [4], following the standardized evaluation protocol of the Planning-oriented Autonomous Driving framework (UniAD) [19], which serves as our teacher model and primary baseline. This ensures that comparisons are fair and directly reflect the impact of our distillation framework rather than differences in data or evaluation settings. The nuScenes dataset consists of 1000 diverse real-world driving scenes collected in Boston and Singapore under varied weather, lighting, and traffic conditions. Each scene includes synchronized 6-camera panoramic RGB imagery (approximately $1600{\times}900$ resolution), annotated 3D bounding boxes, high-definition semantic maps, and ego-vehicle trajectories at 2Hz.

We adopt the official train/validation split of 700/150 scenes to match UniAD [19]'s training protocol. In keeping with our deployment focus, TinyBEV processes *only* monocular camera inputs, excluding LiDAR and radar, to address the constraints of cost-sensitive and sensor-limited platforms such as autonomous shuttles, delivery robots, and drones.

Training supervision follows the same task-specific labels used in UniAD [19]: 3D bounding boxes for detection, future agent positions over a 3-second horizon for forecasting, and ego-vehicle future trajectories for planning. Our supervision combines ground truth labels with teacher-provided soft targets via our multi-stage distillation strategy. We retain the original data preprocessing and augmentations from the UniAD [19] pipeline to ensure direct comparability.

By adhering strictly to the unified nuScenes evaluation protocol, we isolate performance gains to our architectural and distillation contributions. While this work focuses on nuScenes for rigorous comparison with UniAD [19], extending TinyBEV to other large-scale autonomous driving benchmarks and alternative teacher models is an important direction for future research.

### 3.2. Implementation and Training Details

We train **TinyBEV** for 20 epochs on 8 NVIDIA V100 GPUs, using a total batch size of 16 (2 scenes per GPU). Optimization is performed using the AdamW optimizer with an initial learning rate of $2 \times 10^{-4}$ and weight decay of $1 \times 10^{-2}$. The learning rate follows a cosine annealing schedule, with a one-epoch linear warmup at the beginning of training.

The TinyBEV model uses a lightweight ResNet-18 backbone, which is *not* pretrained on ImageNet or any other external dataset. Instead, all weights are initialized randomly. Despite the absence of any large-scale pretraining, our distillation framework enables the student to learn rich spatial-temporal representations by mimicking the teacher's outputs.

The teacher model (UniAD [19]) is frozen throughout training, and its outputs—including intermediate BEV features and task-specific predictions—are precomputed and cached offline. This setup ensures faster training and removes any runtime bottlenecks from computing the teacher's outputs on-the-fly.

To balance the supervised and distillation losses, we empirically set the following loss weights: $\lambda_{feat} = 1.0$, $\lambda_{det} = 0.2$, $\lambda_{mot} = 0.5$, $\lambda_{plan} = 0.5$, and $\lambda_{adapt} = 0.5$. These hyperparameters were selected through grid search on the validation set to ensure fast convergence and optimal task-level performance. Notably, we observed that over-weighting the feature-level distillation led to slower convergence, while under-weighting the output-level distillation caused significant drops in forecasting and planning accuracy.

### 3.3. Evaluation Metrics

To rigorously assess the effectiveness of TinyBEV, we evaluate its performance across three primary tasks in the nuScenes benchmark [4]: 3D object detection, multi-agent motion understanding (via tracking-style metrics), and ego trajectory planning. These tasks are core components of autonomous driving systems, and each is evaluated using task-specific metrics that are standardized and widely adopted in the field.

**Detection.** For the 3D detection task, we report the mean Average Precision (mAP) and the nuScenes Detection Score (NDS). mAP measures the average intersection-over-union (IoU) overlap between predicted and ground-truth 3D bounding boxes across object categories, providing a comprehensive view of object localization and classification accuracy. NDS is a holistic score that aggregates mAP with other metrics, including translation, scale, orientation, velocity, and attribute accuracy. These metrics are crucial for safety-critical systems where precise localization and recognition of dynamic agents are required. A higher mAP or NDS indicates stronger detection capabilities.

**Motion Understanding (Tracking).** Although TinyBEV does not implement an explicit multi-object tracking head, we inherit temporal association capabilities from the UniAD [19] teacher during distillation. As such, we evaluate motion understanding using **AMOTA** (Average Multi-Object Tracking Accuracy) and **AMOTP** (Average Multi-Object Tracking Precision), which capture temporal consistency and spatial precision of agent trajectories over time. This evaluation reflects the model's ability to maintain coherent identities and positions of agents across frames, which is essential for downstream forecasting and planning. We do *not* report minADE here, as our design focuses on distilled spatio-temporal consistency rather than producing multi-modal trajectory hypotheses.

**Planning.** For ego trajectory planning, we assess both the accuracy and safety of the predicted path. Specifically, we compute the L2 distance error at 3 seconds (L2@3s), which reflects how closely the predicted ego trajectory matches the ground-truth route, and the collision rate, which quantifies how often the ego vehicle's predicted path overlaps with other traffic participants. A lower L2@3s error implies better path-following accuracy, while a lower collision rate is essential for ensuring safe motion in complex driving scenarios.

**Overall Evaluation.** We summarize these metrics in Table 2, which compares different variants of TinyBEV (S0–S3) with the teacher model UniAD [19]. The re- sults clearly show that TinyBEV with joint distillation (S3) nearly matches the performance of the teacher in all evaluated tasks. Notably, S3 achieves an mAP of 39.0 and AMOTA of 34.0, just slightly below UniAD [19]'s 41.0 and 36.0, respectively. In terms of planning, TinyBEV (S3) reaches a L2@3s of 1.08 m and collision rate of 0.32%, nearly indistinguishable from the teacher (1.03 m and 0.31%). These results demonstrate the efficacy of our cross-task, cross-level distillation pipeline in enabling a compact model to perform robustly in real-world autonomy settings.

### 3.4. Experimental Variants

To evaluate the effectiveness of different knowledge distillation strategies, we conduct a comprehensive ablation study across four student configurations. These variants are designed to isolate and understand the impact of *feature-level* and *output-level* supervision from the teacher model (UniAD [19]). The three evaluated tasks are 3D object detection, motion understanding (evaluated via tracking-style metrics), and ego trajectory planning. While TinyBEV does not include an explicit tracking head, it inherits temporal association capabilities from UniAD [19] through distillation, enabling us to report AMOTA as a proxy for multi-agent motion consistency.

- **S0 (No Distillation):** Baseline model trained solely using ground-truth (GT) supervision for all tasks—detection, motion understanding, and planning—without any teacher guidance. This setup provides a lower bound on performance, showing how far a compact BEV model can go without leveraging teacher knowledge.
- **S1 (Feature-Level Distillation):** Student receives intermediate BEV feature maps distilled from the teacher. This encourages learning more expressive and structured spatial-temporal representations, which is especially beneficial in multi-task settings where perception and prediction are tightly coupled.
- **S2 (Output-Level Distillation):** Student is supervised using the teacher's final outputs—detection logits, agent trajectory predictions (for motion understanding), and ego planning outputs. This high-level logit matching encourages the student to replicate the teacher's decision-making behavior, improving task-specific predictions.
- **S3 (Full Distillation):** Proposed configuration combining both feature-level and output-level distillation. This holistic learning signal aligns intermediate representations while also transferring the teacher's final outputs, leading to the most consistent gains across all tasks.

These variants are outlined in Table 3, which summarizes the knowledge transfer pathways used in each setup.

### 3.5. Quantitative Results

Table 2 presents a consolidated multi-metric evaluation of TinyBEV student variants (S0–S3) and the UniAD [19] teacher on the nuScenes validation set. We report detection (mAP, NDS), tracking (AMOTA, AMOTP), planning (L2 error at 3s, minADE), safety (ego-agent collision rate), and efficiency (FPS, parameters). All numbers are taken from our best evaluation runs for consistency.

| Method | mAP↑ | NDS↑ | AMOTA↑ | AMOTP↓ | L2@3s↓ | minADE↓ | Collision↓ | FPS↑ |
|---|---|---|---|---|---|---|---|---|
| Teacher (UniAD [19]) | 41.0 | 35.0 | 36.0 | 1.50 | 1.03 | 0.70 | 0.31 | 2 |
| TinyBEV (S0) | 31.0 | 30.0 | 28.0 | 2.00 | 1.43 | 1.00 | 0.48 | 11 |
| TinyBEV (S1) | 38.0 | 32.1 | 30.0 | 1.75 | 1.40 | 0.85 | 0.45 | 11 |
| TinyBEV (S2) | 31.0 | 30.5 | 29.0 | 1.90 | 1.22 | 0.82 | 0.39 | 11 |
| TinyBEV (S3) | 39.0 | 34.2 | 34.0 | 1.60 | 1.08 | 0.78 | 0.32 | 11 |

Table 2. **Unified evaluation on nuScenes.** S3 achieves the best overall trade-off, approaching UniAD [19]'s performance while running over 5× faster with 78% fewer parameters (28M vs. 125M).

**Key findings:**
- **Detection:** S3 reaches 39.0 mAP—close to the teacher's 41.0—and improves substantially over the non-distilled baseline (S0: 31.0).
- **Tracking:** Although TinyBEV lacks a dedicated tracking head, AMOTA is computed using the standard nuScenes association protocol. Distillation of temporally consistent BEV features from UniAD [19] enables S3 to maintain strong temporal coherence (34.0 vs. 36.0 for UniAD [19]).
- **Planning:** S3 attains low planning error (L2@3s: 1.08, minADE: 0.78) and a collision rate of 0.32%, closely matching the teacher's 0.31%.
- **Efficiency:** All TinyBEV variants run at 11 FPS—over 5× faster than UniAD [19]—while using 78% fewer parameters.

### 3.6. Ablation Results

To assess the contribution of each distillation strategy, we evaluate the four configurations defined in Table 3, isolating the effects of feature-level, output-level, and combined knowledge transfer from the UniAD [19] teacher.

| Experiment | Description |
|---|---|
| S0 | Baseline TinyBEV trained only with ground-truth supervision (no teacher). |
| S1 | *Feature KD*: Distills intermediate BEV features from the teacher. |
| S2 | *Output KD*: Distills final prediction logits (detection, forecasting, planning) from the teacher. |
| S3 | *Full KD*: Combines feature- and output-level distillation. |

Table 3. **Distillation configurations.** S0: no KD; S1: feature-level only; S2: output-level only; S3: combined.

Figure 2 summarizes the performance of each variant on detection (mAP), forecasting (minADE), planning (L2@3s), and safety (collision rate).

**Key observations:**
- **Feature KD (S1)** yields substantial planning and forecasting improvements over S0, benefiting from richer spatial-temporal features.
- **Output KD (S2)** produces sharper task predictions, with strong forecasting gains, but slightly smaller planning improvements compared to S1.
- **Full KD (S3)** consistently achieves the best results across all metrics, confirming that aligning both intermediate representations and final outputs provides complementary benefits.

| Variant | mAP↑ | minADE↓ | L2@3s↓ | Collision↓ |
|---|---|---|---|---|
| S0 | 31.0 | 1.00 | 1.43 | 0.48 |
| S1 | 38.0 | 0.85 | 1.30 | 0.45 |
| S2 | 31.0 | 0.82 | 1.22 | 0.39 |
| S3 | **39.0** | **0.78** | **1.08** | **0.32** |

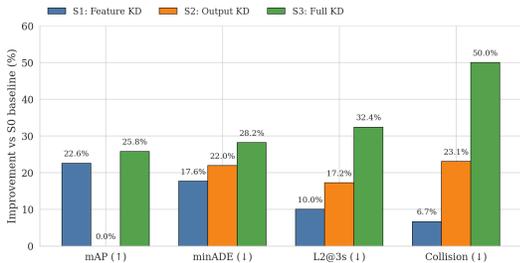

Figure 2. **Ablation study.** Top: raw performance metrics. Bottom: relative change vs S0. For mAP/minADE/L2@3s/Collision, improvement is computed as $(M_{\text{variant}} - M_{S0})/M_{S0} \times 100$ for higher-is-better metrics (mAP), and $(M_{S0} - M_{\text{variant}})/M_{S0} \times 100$ for lower-is-better metrics (minADE, L2@3s, Collision).

### 3.7. Qualitative Results

Figure 3 compares the outputs of the UniAD [19] teacher and our best-performing TinyBEV (**S3**) student on the same driving scene. Each row displays synchronized outputs from six camera views alongside the bird's-eye-view (BEV) projection. The BEV maps visualize 3D object detections (colored boxes), motion forecasts (colored trajectories), and the ego-vehicle's planned path (orange dotted line), with temporal color encoding for direction and speed.

**Observations:**
- Predictions are visually consistent with the teacher, confirming that both mid-level (feature) and high-level (output) distillation contribute to preserving spatial detail and motion structure.
- TinyBEV maintains accurate planning and collision avoidance behavior, even under moderate occlusions.
- Minor deviations in long-range agent detection do not significantly affect downstream planning quality.

## 4. Limitations and Future Work

While **TinyBEV** achieves competitive performance in real-time, full-stack autonomous driving via multi-stage distillation from the UniAD teacher [19], several limitations suggest avenues for future research.

**Limited Sensor Modalities.** TinyBEV is a *camera-only* system, omitting LiDAR and radar that many high-performance systems exploit for robustness [34, 35, 50, 51]. This choice maximizes efficiency and deployment flexibility, but may reduce resilience in adverse conditions (e.g., weather/lighting) discussed in autonomy surveys [1, 11, 17, 21]. A pragmatic direction is *lightweight multi-sensor fusion* that preserves real-time operation [34, 51].

**Teacher Dependency and Distillation Scope.** Our approach assumes access to a strong teacher producing dense supervisory signals [25, 27, 41, 47, 48, 54]. This can constrain applicability under domain shift or where such teachers are unavailable [1, 17]. Future work may explore *teacher-lite* or *teacher-free* variants grounded in the same cross-modal/task distillation principles [25, 41] and broader representation learning insights summarized by recent surveys [1, 21].

**Dataset and Scenario Coverage.** Our evaluation centers on nuScenes [4]. Although widely used, it cannot represent all rare edge cases and scene geometries. Extending to complementary benchmarks (e.g., KITTI [16]) and stress-testing under distribution shifts recommended by surveys [1, 21] will better quantify generalization and failure modes.

**Planning and Complex Behavior Modeling.** While TinyBEV closely matches the teacher on planning/forecasting, richer long-horizon reasoning remains challenging [5, 19]. Promising directions include *temporal aggregation* and *sequence modeling* in BEV [22, 33, 38], as well as tighter perception–map–motion coupling [10, 32, 44, 53].

**Robustness to Teacher Quality.** How student performance degrades with weak/noisy teachers remains underexplored. Robust KD formulations and noise-tolerant objectives in 3D/BEV settings [27, 47, 54] merit deeper study.

**Online Adaptation and Continual Learning.** Our training is offline with a fixed teacher. Surveys emphasize the importance of adaptation and continual learning for autonomy in the wild [1, 17]. Incorporating lightweight on-device updates without sacrificing safety/latency is a valuable next step.

**Planned Extensions.** We plan to: (1) integrate lightweight radar/camera fusion [34, 37, 51]; (2) relax teacher dependence via cross-task/modal distillation advances [25, 41]; (3) broaden evaluation beyond nuScenes [4] to KITTI [16] and additional settings highlighted in surveys [1, 21]; (4) study online/continual adaptation for robustness [17]; and (5) further optimize efficiency with compression and deployment-aware design [18, 43, 55].

## 5. Conclusion

We introduced **TinyBEV**, a compact, camera-only full-stack model distilled from the large UniAD [19] teacher [19]. By combining *feature-level* and *output-level* supervision within a unified BEV-centric architecture [20, 28, 31, 39, 49], TinyBEV attains near-teacher accuracy

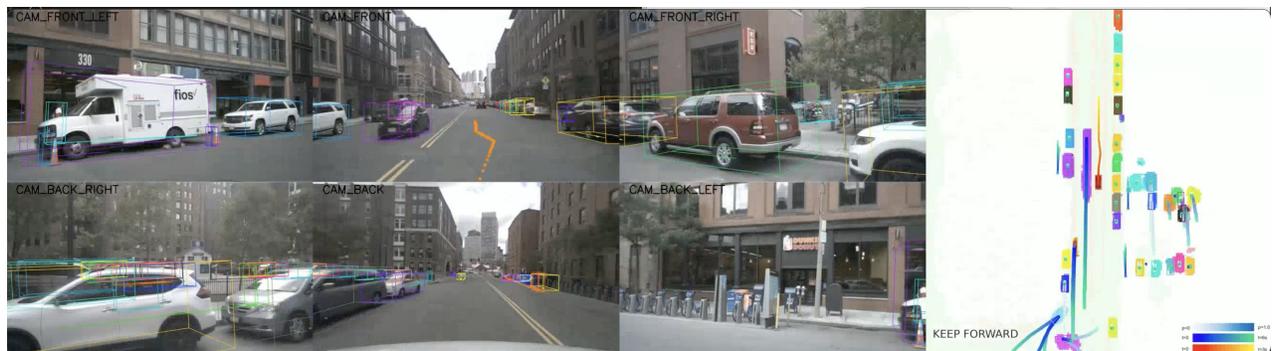

(a) **UniAD [19]. (Teacher)**: High-capacity transformer-based model producing dense and temporally consistent predictions without ground-truth labels. Serves as the distillation source for TinyBEV.

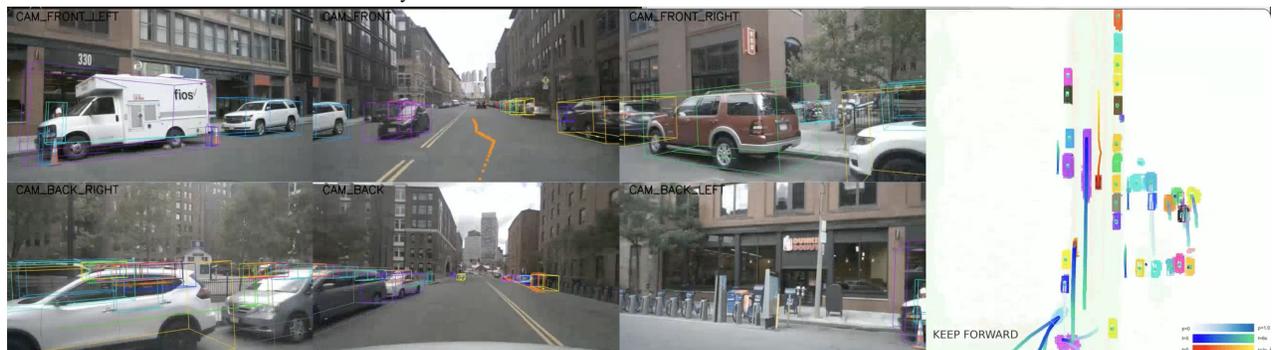

(b) **TinyBEV (Student, S3)**: Compact camera-only model achieving comparable spatial and temporal precision while running at 11 FPS with 78% fewer parameters.

Figure 3. **Qualitative comparison between teacher and student predictions.** TinyBEV closely matches UniAD [19]. in 3D object localization, multi-agent trajectory forecasting, and goal-directed planning, despite being over 4× faster. Minor differences appear in far-range detections (circled in the BEV for visibility). These results align with the quantitative improvements reported in Table **??** and demonstrate that distillation transfers both scene understanding and safe driving intent.

across detection, forecasting, and planning on nuScenes [4], while running significantly faster and with far fewer parameters.

Our ablations underline that joint feature/output distillation and temporal BEV modeling [22, 38] are complementary, and our qualitative analyses echo structured scene understanding consistent with full-stack systems [5, 10, 44]. Looking forward, combining distillation advances [25, 27, 41, 47] with efficiency techniques [18, 43, 55] and broader evaluation [1, 16, 21] promises robust, scalable autonomy on resource-constrained platforms.